\documentclass[10pt,conference,a4paper,twocolumn] {IEEEtran}

\usepackage[utf8]{inputenc}
\usepackage{algorithm} 
\usepackage{algpseudocode}
\usepackage{array}
\usepackage{graphicx}
\usepackage{epic}
\usepackage{figsize}
\usepackage{float}
\usepackage{hyperref}
\usepackage[section]{placeins}
\usepackage{cite}
\usepackage{amsmath}
\usepackage{amssymb}
\usepackage{amsfonts}
\graphicspath{ {./images/} }

\newcommand{\email}[1]{{\fontfamily{cmtt}\selectfont#1\normalfont}}
\newcolumntype{P}[1]{>{\centering\arraybackslash}p{#1}}

\IEEEoverridecommandlockouts

\title{Model-Free Deep Reinforcement Learning in Software-Defined Networks}
\author{\IEEEauthorblockN{Luke Borchjes\IEEEauthorrefmark{1},Clement Nyirenda\IEEEauthorrefmark{2},Louise Leenen\IEEEauthorrefmark{3}}
\IEEEauthorblockA{\emph{Computer Science Department, University of the Western Cape}\\
$^1$\email{ldborchjes@gmail.com, 3647745@myuwc.ac.za}\\
$^2$\email{cnyirenda@uwc.ac.za}\\
$^3$\email{lleenen@uwc.ac.za}\\
\emph{South Africa}\\
}}

\begin{document}

\maketitle
\begin{abstract}
This paper compares two deep reinforcement learning approaches for cyber security in software defined networking. Neural Episodic Control to Deep Q-Network has been implemented and compared with that of Double Deep Q-Networks. The two algorithms are implemented in a format similar to that of a zero-sum game. A two-tailed T-test analysis is done on the two game results containing the amount of turns taken for the defender to win. Another comparison is done on the game scores of the agents in the respective games. The analysis is done to determine which algorithm is the best in game performer and whether there is a significant difference between them, demonstrating if one would have greater preference over the other. It was found that there is no significant statistical difference between the two approaches.
\end{abstract}
\vspace{0.2cm}
\section{INTRODUCTION}
Most cyber security approaches are model-based and lack scalability because they are sample inefficient \cite{Nguyen_2021}. Literature has also shown that there is a lack of networking models featuring the next generation of networks, such as Software Defined Networks (SDNs) \cite{Yi_Han_et._al.}, \cite{Nguyen_2021}. Interest in using deep reinforcement learning (DRL) has grown significantly due to its versatility; One example for DRL approaches is the Double Deep Q-Learning algorithm \cite{Hasselt_et._al._2015}. Double Deep Q-Networks (DDQN) emanates from the combination of Double Q-Learning and Deep Q-Networks \cite{Hasselt_et._al._2015}. This algorithm is aimed at reducing the number of over estimations by splitting the maximum operation of the target function into two Q-value functions, action selection and action evaluation. The target network of the Deep Q-Network (DQN) architecture replaces the action evaluation function, giving the resulting DDQN algorithm.

A recent modification to Deep Q-Networks has introduced Neural Episodic Control (NEC) into the algorithm \cite{nishio2018faster}. Neural Episodic Control (NEC) to Deep Q-Network has an implementation of similar structure to DDQN, although instead of double Q-Learning with a DQN it makes use of neural episodic control and a DQN \cite{Hasselt_et._al._2015} Neural episodic control replaces the evaluation network. NEC has greater performance scores in earlier iterations, whereas DQN outperforms it in latter iterations \cite{Alexander_Pritzel_et_al_2017}. It is this combination that gives it the shorthand names NEC2DQN and N2D \cite{Alexander_Pritzel_et_al_2017}. This paper is aimed at finding out whether these algorithms are capable of managing a network and isolating machines that are infected and prevent further propagation throughout the network, preventing the intruder from gaining access to the critical server within the context of software defined networks (SDN).   

The rest of the paper is organised as follows. Section I presents the fundamentals of software defined networking (SDN), including the definition, motivation as well as some literature on SDN. Section II presents the concept of deep reinforcement learning, focusing on the chosen algorithms, their definition, motivation as well as the associated literature. Section III presents the methodology, which includes the implementation of the environment as well as agent use of the algorithms. The results are presented and analysed in Section IV. The discussion and future work are presented in Sections V and VI respectively. 

\section{BRIEF OVERVIEW ON SOFTWARE DEFINED NETWORKING}
Software Defined Networking (SDN) is a three layer network architecture that has been developed and put in practice as early as of 2013. SDN is composed of the three layers: (1) application layer made up of applications, delivering services and communicating their network requirements to the controller by means of northbound APIs; (2) Control layer, hosting the SDN controller, translating requirements into low-level controls sent through southbound API's to the infrastructure layer; (3) infrastructure layer, consisting of network switches \cite{Yi_Han_et._al.}. The major advantage of SDN is that it separates network control and forwarding functions, allowing the controller to be programmable to perform various application services and tasks \cite{Yi_Han_et._al.}. Consequently, network resources can be conveniently managed, configured and optimised using the standardised protocols. Due to its architecture there has been a good variety of available open-source SDN controller platforms/frameworks, a few examples being OpenDayLight, RYU, NOX/POX and Open vSwitch \cite{Yi_Han_et._al.}.

In this paper, RYU was chosen as it is recommended for quick prototyping, being consistently updated and well built with Python. RYU is a component-based software defined networking framework, providing APIs to create, manage or control networks \cite{ryu}. This controller is paired with MiniNet, a system framework that creates a realistic virtual network, with a real kernel, switch and application code that can be run on a single machine either native, through cloud services or on a virtual machine (VM) \cite{mininet}. Reinforcement learning in SDN has been demonstrated as a fitting tool for various purposes as demonstrated by \cite{Yi_Han_et._al.}. A good example is the implementation of intelligent routing based on reinforcement learning for software defined networking by \cite{intelligent_routing}. A challenge SDN faces stems from dynamic traffic patterns, and requires frequent network reconfiguration \cite{Software_Defined_Networking}. Consequently, reinforcement learning presents itself as an ideal tool for this task.

\section{DEEP REINFORCEMENT LEARNING FUNDAMENTALS}

\subsection{Double Deep Q-Learning}
Q-Learning is a reinforcement learning algorithm proposed to optimally solve Markov Decision Processes (MDPs) \cite{Otterlo2012MArkovDP}. An MDP is defined as (S, A, P, R, GAMMA); $S$ is the states, $A$ is the set of actions doable in the environment, $P$ the state transition property,  $R$ the reward and finally Gamma ($\gamma$) is the discount factor \cite{Otterlo2012MArkovDP}. However in stochastic MDPs Q-Learning performs poorly due to the large over estimations of the action-values \cite{Van_Hasselt_Hado_2010}. The algorithm contains the $Q$ function that calculates the quality of a state-action combination; $Q : S \times A \rightarrow \mathbb{R}$. The core of the algorithm is a Bellman equation as a simple value iteration update, using the weighted average of the old value and the new information by using
\begin{equation} \label{q_new}
Q_{new}(s_{t}, a_{t})\leftarrow Q(s_{t},a_{t})+(r_{t}+ max_{a}Q(s_{t+1},a)-Q(s_{t}, a_{t})).
\end{equation}

The $Q$ function works by initialising it to any possibly arbitrary fixed value. Then at each time $t$, agent is in state $s_t$ and selects an action $a_t$, observes a reward $r_t$ and enters a new state $s_{t+1}$, then updates $Q$. The action selected is linked to the highest expected value. In doing so, the obvious method to obtain it, is by approximation of the value by means of the maximal estimator ($max_{a}Q(s_{t+1},a)$) as seen in the temporal difference $(r_{t}+ max_{a}Q(s_{t+1},a)-Q(s_{t}, a_{t})$ in the $Q$ function in Eq. (1). 

The approximation of the value of the next state is done by maximising over the estimated action values in that state. To solve the overestimation of action-values the algorithm Double Q-Learning is proposed. Double Q-Learning is the implementation of two $Q$ functions: $Q_{A}$ and $Q_{B}$. Each $Q$ function is updated from the other’s next state \cite{Van_Hasselt_Hado_2010} by using
\begin{equation} \label{q_a_1}
Q_{A} \leftarrow Q_{A}+(r_{t}+max_{a}Q_{B}-Q_{A}),
\end{equation}
where $Q_{A}=Q_{A}(s,a)$, $Q_{B}=Q_{B}(s',a^*)$, and $a^*=argmax_{a}Q_{A}(s', a)$.
\begin{equation} \label{q_b_1}
Q_{B} \leftarrow Q_{B}+(r_{t}+max_{a}Q_{A}-Q_{B}),
\end{equation}
where $Q_{B}=Q_{B}(s,a)$, $Q_{A}=Q_{A}(s',b^*)$, and $b^*=argmax_{a}Q_{B}(s', a)$.

In this paper a modified version of the algorithm proposed by Hasselt et. al. \cite{Hasselt_et._al._2015} is employed; Algorithm 1 shows the implementation of this algorithm.

\begin{algorithm}
    \caption{Double Q-Learning (Hasselt et al., 2015)}
    \begin{algorithmic}[1]
        \State Initialise networks ${Q}_{\theta}$ and ${Q}_{\theta'}$
        \State Initialise replay buffer D
        \State Initialise $\tau << 1$
        \For{each iteration}
            \For{each environment step}
                \State Observe state $s_{t}$ and select action $a_{t} \sim \pi(a_{t}, s_{t})$
                \State Execute $a_{t}$
                \State Observe next state $s_{t+1}$ and reward $r_{t} = {R}(s_{t}, a_{t})$
                \State Store $(s_{t}, a_{t}, r_{t}, s_{t+1})$ in replay buffer $D$
            \EndFor
            \For{each update state}
                \State sample $e_t = (s_{t}, a_{t}, r_{t}, s_{t+1}) ~ {D}$
                \State Compute target Q value using (2, 3)
                \State Perform gradient descent step on:
                \State $(Q*(s_{t}, a_{t})-Q_{\theta}(s_{t}, a_{t}))^{2}$
                \State Update target network parameters:
                \State $\theta' \leftarrow \tau * \theta + (1 - \tau) * \theta'$
            \EndFor
        \EndFor
    \end{algorithmic}
\end{algorithm}
\label{Algorithm1}

\subsection{Neural Episodic Control to Deep Q-Learning}
Neural Episodic Control is an algorithm proposed by Pritzel et. al. in \cite{Alexander_Pritzel_et_al_2017}. NEC is able to execute successful strategies as soon as they are experienced, instead of waiting for optimization to be done, such as stochastic gradient descent as with Deep Q-Networks in Q-Learning. The algorithm has three parts, a convolutional neural network for the processing of images and pictures (used to create a state embedding), a set of memory modules (one per action) and a final network for converting action readouts into $Q(s, a)$ \cite{Alexander_Pritzel_et_al_2017}. The memory module mentioned is referred to as a differential neural dictionary (DND) because each memory module contains two dynamically sized arrays of vectors, $K_a$ and $V_a$. Memory module $M_a$ exists for each action $a \in A$ such that $M_a = (K_a, V_a)$.  For array $K_a$, key $h$, for $h \in K_a$, is produced by inputting pixel state $s$ into CNN and is used to lookup a value from DND $M_a$ yielding the weights $w$. The key $h$ is also described as the embedding vector. Array $V_a$ holds the target values $v \in V_a$ for each action. The DND has two function operations: lookup and write. When doing a lookup using key $h$ it returns an output $o$, the weighted sum of the values in the DND, defined by
\begin{equation}
    o = \sum_i w_i v_i,
\end{equation}
where $w_i= \frac{k(h,h_i)}{\sum_j k(h, h_i)}$ and 
\begin{equation}
  k(h,h_i) = \frac{1}{||h-h_i||^2 _2 + \delta}.
  \label{kh_i}
\end{equation}

Weight $w_i$ is given by normalised kernels between vectors $h$ and $h_i$, the lookup key and the corresponding key in memory respectively. The kernel function used for $h$ and $h_i$ is Eq. \ref{kh_i} \cite{nishio2018faster}. The write operation simply appends the keys and values to their corresponding arrays $K_a$ and $V_a$. Should a key-value pair already exist it is updated. The values in the DND are in turn the corresponding $Q$ values to the state originally having resulted in the key-value pair to be written to memory. Therefore producing an estimate of $Q(s, a)$ for  any single given action $a$ \cite{nishio2018faster}. To update the values in DND N-step Q-learning is applied as in \cite{mnih_et_al_2016}, thus the N-step Q-Value estimate is then
\begin{equation}
    Q^{(N)}(s_t,a) = \sum^{N-1} _{j=0}  \gamma^j r_{t+j}+ \gamma^N max_{a'} Q(s_{t+N},a')
\end{equation}
Thus the value in array $V_a$ is $Q_i$ for:
\begin{equation}
Q_i \leftarrow Q_i+ (Q^{(N)}(s,a) - Q_i)
\end{equation}

Neural Episodic Control, however, requires large memory and a lot of calculation time with its computational space and time \cite{Alexander_Pritzel_et_al_2017}. Thus in \cite{nishio2018faster} a  solution is proposed in the form of the algorithm called Neural Episodic Control to a Deep Q-Network, or shorthand NEC2DQN (N2D). This algorithm is created by supplementing NEC with the simplest Deep Q-learning network. This is possible because with both NEC and DQN there is always one action-value Q*, thus it is possible for them to converge on the same Q*. Therefore the Q-value of one algorithm can be taken as the target value, making it easier to converge towards a better target value, approaching Q* faster\cite{nishio2018faster}. Similarly this is seen in the Double DQN algorithm mentioned previously. By setting, $Q_{A}(s, a)=Q_{DQN}(s, a), Q_{B}(s, a)=Q_{NEC}(s, a)$ we can modify (2) and (3) such that
\begin{equation}
Q_{DQN}(s, a) \leftarrow Q_{DQN}(s, a)+(r+Q_{NEC}(s',a^*)-Q_{DQN}(s, a)),
\end{equation}
and 
\begin{equation}
Q_{NEC}(s, a) \leftarrow Q_{NEC}(s, a)+(r+Q_{DQN}(s',b^*)-Q_{NEC}(s, a)).
\end{equation}
Since NEC uses N-step learning we adopt and rewrite the equations as; 
\begin{equation}
Q_{DQN}(s, a) \leftarrow Q_{DQN}(s, a)+(r+Q_{N2D}^{(N)}(s',a^*)-Q_{DQN}(s, a))
\end{equation}
\begin{equation}
Q_{NEC}(s, a) \leftarrow Q_{NEC}(s, a)+(r+Q_{N2D}^{(N)}(s',b^*)-Q_{NEC}(s, a)),
\end{equation}
where $Q_{N2D}^{(N)}$ is defined as 
\begin{equation}
Q_{N2D}(s_t, a)=\lambda(t)Q_{NEC}(s_t, a)+(1-\lambda(t))Q_{DQN}(s_t,a).
\end{equation}

Algorithm 2 shows the implementation of Neural Episodic Control to Deep Q-Learning. Initially, a Convolutional Neural Network (CNN) is used because the input states contained pixels that were processed. Algorithm 2 was then modified such that instead of using a CNN to getting the state embedding $h$, an embedding neural network was used. The modification was necessary because the state provided is a 1-dimensional array of length 80, containing 1's and 0's.

\begin{algorithm}
    \caption{Neural Episodic Control to Deep Q-Learning (Nishio et al., 2018)}
    \begin{algorithmic}[1]
        \State Initialise the number of time steps
        \State Initialise the change step CS for $\lambda$(t)
        \State Initialise replay memory D to capacity ${C}_{D}$
        \State Initialise replay memory E to capacity ${C}_{E}$
        \State Initialise DND memory ${M}_{a}$ capacity ${C}_{M_a}$
        \State Initialise action-value function ${Q}_{NEC}$ and ${Q}_{DQN}$
        \For {each episode}
            \For {$t = 1, 2, ..., T$}
                \State Receive observation $s_{t}$ from environment
                \If {$ {t}$ $mod(2)$}
                    \State Receive $Q_{DQN}(s_{t},a)$ 
                    \If {$S < CS$ }(use NEC)
                        \State Receive embedding $h$ and $Q_{NEC}(s_{t},a)$. 
                    \Else
                        \State Set $Q_{NEC}(s_{t},a)$ to free values. 
                    \EndIf 
                    \State Calculate $Q_{N2D}(s_{t},a)$
                    \State $a_{t} \leftarrow \epsilon-greedy$ policy on $Q_{N2D}(s_{t},a)$
                    \State Take action $a_{t}$, receive reward $r_{t}$. 
                    \State Append ($s_{t},a_{t},r_{t}$) to $G$. 
                    \State Train on a random minibatch from $D$.
                    \State $S \leftarrow S+1$
                \EndIf
                \If {not ${t}$ $mod(2)$}
                    \State Receive $Q_{DQN}(s_{t},a)$
                    \State $a_{t} \leftarrow \epsilon-greedy$ policy on $Q_{DQN}(s_{t},a)$
                    \State Take action $a_{t}$, receive reward $r_{t}$. 
                    \State Append $(s_t,a_t,r_t,s_{t+1})$ to $E$
                    \State Train on random minibatch from $E$
                \EndIf
                \For {$t = 1, 2, ..., T$}
                    \State Calculate $y_{t}$
                    \State Append ($s_{t},a_{t},r_{t}$) to $G$
                    \If {$S < CS$}
                        \State Append ($h_{t},y_{t}$) to $M_{a}$
                    \EndIf
                \EndFor
            \EndFor
        \EndFor
    \end{algorithmic}
\end{algorithm}

The algorithms process information, learn and make the decision that best maximises the return reward for a chosen action. The algorithms use Q-values to determine the state-action value pair that produces the action which maximises the reward return. They rely on the experiences stored in a memory buffer. To first get these experiences exploration is introduced, where an action is selected at random from the action space available. Memory batches are sampled and used in the training of the model using the algorithm that is built using the equations explored. As the agent using the model progresses and adapts, exploitation is used. Exploitation is when the model makes a prediction based on what it's learnt, instead of using random exploration. The shift is gradual and in this case is implemented using the epsilon-greedy method, in order to balance exploration vs exploitation.

\section{METHODOLOGY}
For the execution of the simulation, all the work was done in Ubuntu 20.04 with the programming language of choice Python3 alongside the Tensorflow2, Cuda and Mininet frameworks as well as the RYU network controller. The hardware used to run the simulations was an Nvidia RTX 3080Ti, Intel i7 10700F, with 64 Gigabytes of DDR4 RAM. 

The algorithms presented in the previous section are implemented in a way that they take in a state, then either by exploration or exploitation, make a decision on an optimal action. That action is then implemented into the environment and in return a new environment state and reward is returned. As illustrated in Fig 1., two different agent algorithms are implemented on a turn based system. 

\begin{figure}[h]
    \centering
    \includegraphics[scale=0.4]{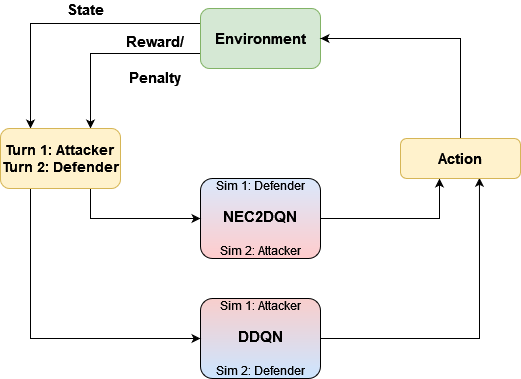}
    \caption{Block diagram of Deep Reinforcement Learning implementation}
    \label{fig:mesh1}
\end{figure}

The agents using these algorithms play two different roles alternatively. During runs of game 1 the N2D algorithm is used by the defender and DDQN by the attacker. During the runs of game 2 the N2D algorithm is used by the attacker and DDQN is used by the defender. As illustrated in Fig. 2, the environment makes use of a star topology, with four switches connected to a central router. Each switch connects to a series of hosts that forms its own subnet. In total there are 4 four subnets with subnet 1 having 6 hosts, subnet 2 with 8 hosts as well as subnets 3 and 4 with 9 hosts each. The environment is hosted on Ubuntu running MiniNet and RYU natively. 

\begin{figure}[h]
    \centering
    \includegraphics[scale=0.4]{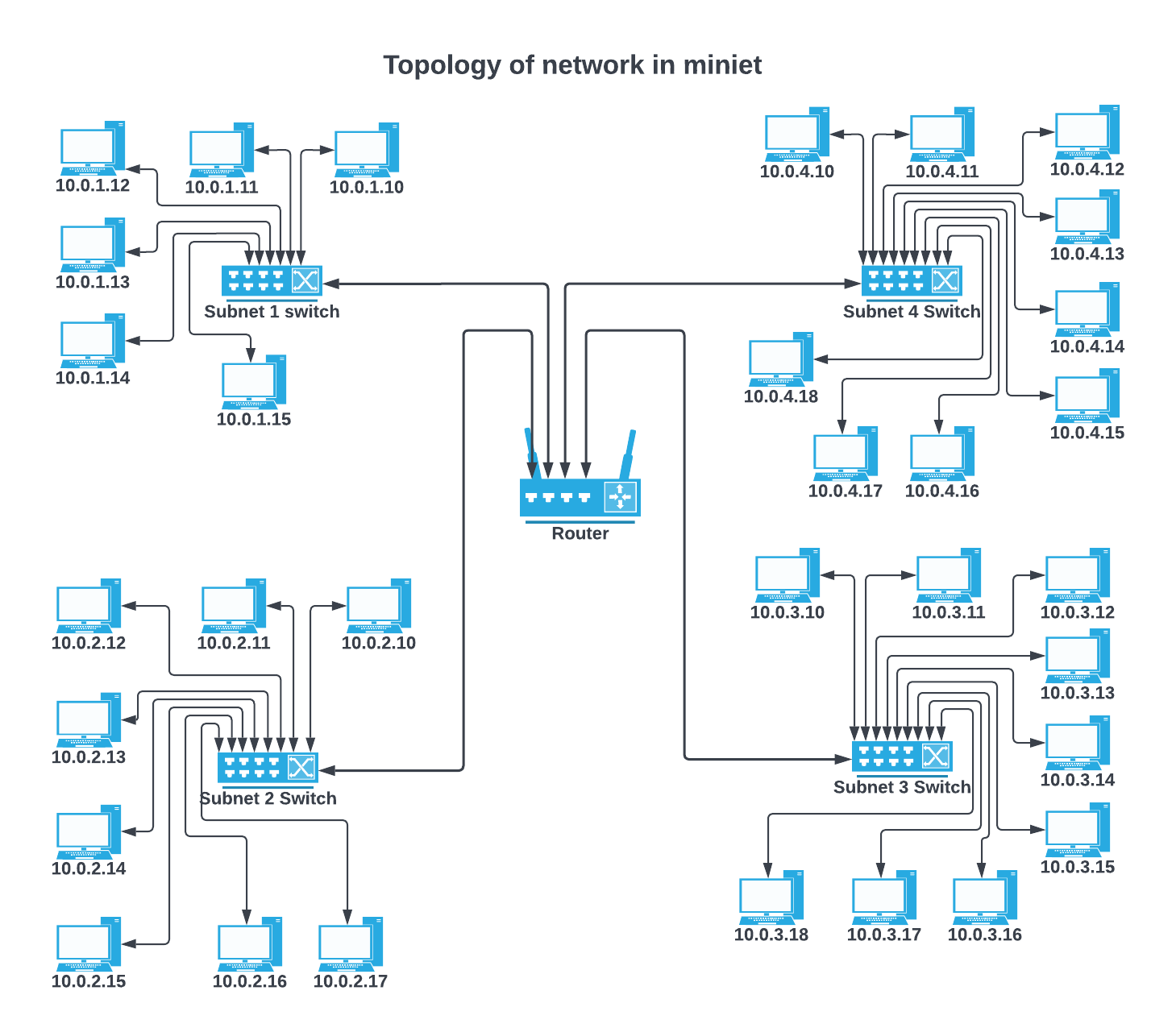}
    \caption{Star topology implemented in MiniNet}
    \label{fig:mesh1}
\end{figure}

Network restrictions and routes were set in place within the SDN using MiniNet, such that only certain hosts were visible to others. Fig. 3 illustrates the network seen by the agents as well as communication routes. In Fig. 3 hosts that are coloured red are those that are initially compromised where as blue indicates the critical server location and the yellow indicates those which are ordinary hosts.

\begin{figure}[h]
    \centering
    \includegraphics[scale=0.45]{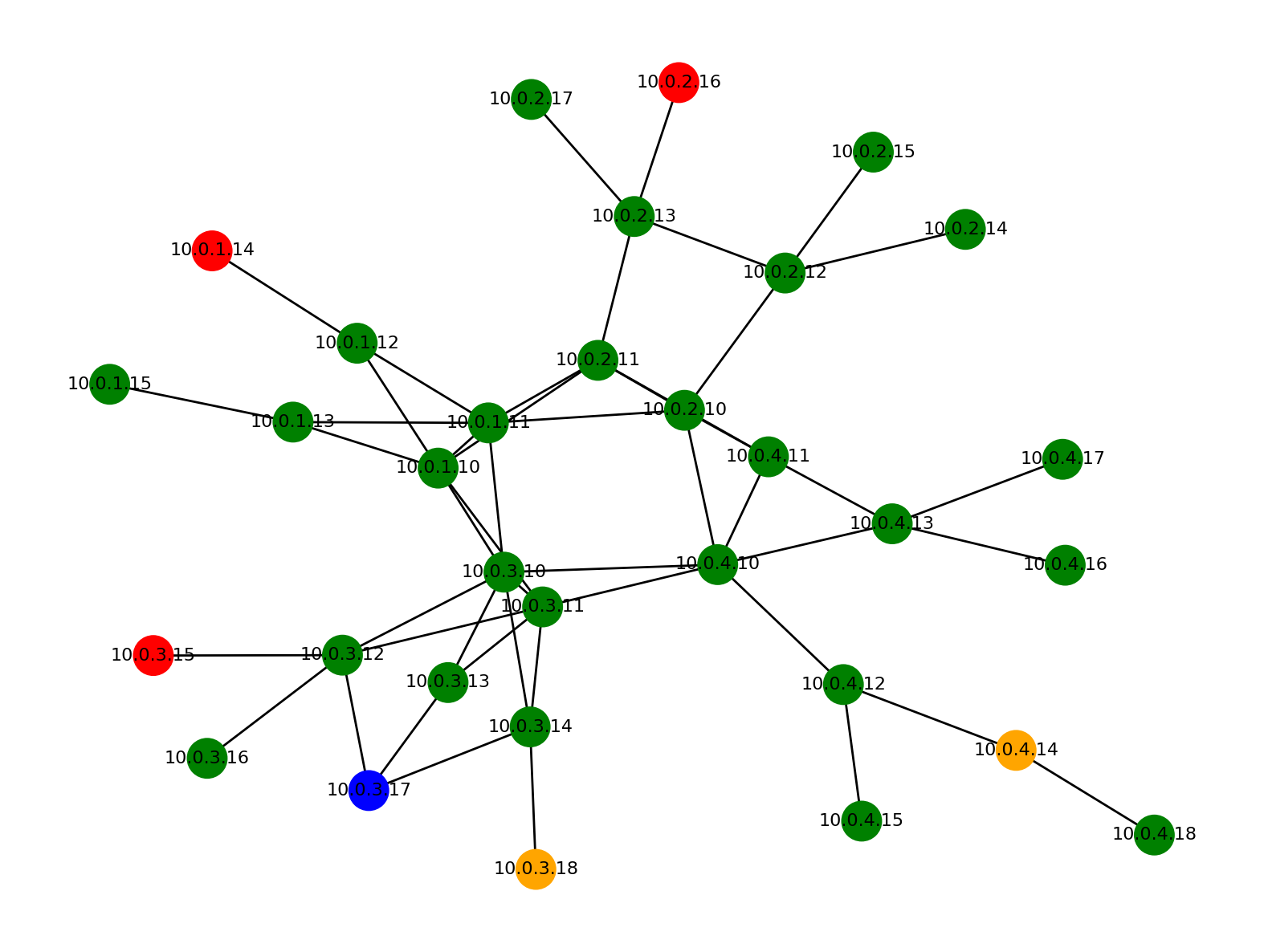}
    \caption{Topology showing the allowed visibility and communication links}
    \label{fig:mesh1}
\end{figure}

The flag system commonly used in hackathon competitions has been implemented into the game. Depending on the role the agent had, they would have a different set of actions as described with different win conditions. For the attacking agent to win, it would have to successfully complete its goal by compromising the critical server, or drain the defender of points, such that its score is greater than the defenders. The attacker can only compromise one host a turn, by injecting a flag into that compromised host. The defending agent wins by completely blocking out the attacker, by isolating and patching all hacked hosts, or by preventing further spread and thus isolating the remaining network from infected sections, or by outlasting the attackers assault and ending the game with a greater score.

The observed state that the agents see is a numpy array of length 80, containing 1’s and 0’s, representing the hosts and links. For the hosts 1 represents not compromised and 0 represents compromised. For the links in the network, the value of 1 represents active and 0 represents inactive. The zero sum game format was chosen when structuring the format of the game loop. For scoring the defender will start with the maximum possible score. The attacker will start at 0. As the attacker gains ground, the defender loses points and the attacker gains points. As the defender regains lost ground then the defender regains those lost points and the attacker loses them. Initially a game ran at a cap of 1,000,000 turns, giving each agent 500,000 turns each. Thus for 10 game runs each agent had a total of 5 million steps. When running at 500,000 turns per agent it was rare that a game run exceeded 50,000 turns, thus the turns were capped at 50,000. Taking this into consideration, later, 10 game runs were done at 25,000 turns per agent. Thus each agent got a total of 250,000 steps. Two games are done where the agents had alternated their roles, implying in game 1 the agent using the N2D algorithm is the defender and in game 2 is the attacker. The attacking agent uses the DDQN algorithm in game 1, and in game 2 is the defending agent. In all games the attacker goes first.

\subsection{Agents}
Two agents are implemented. An agent is classified as red when it has the role of the attacker and blue when it has the role of the defender. An agent is an entity trained to make the most optimal decision by means of the desired algorithm implemented. Both these agents are each implementing a different type of deep reinforcement learning algorithm. The DDQN agent uses the Double Deep Q-Network algorithm proposed in \cite{Hasselt_et._al._2015}, as presented in Algorithm 1. The N2D agent uses the NEC2DQN algorithm as proposed by in \cite{nishio2018faster} and illustrated in Algorithm 2.

The agents alternate in the assignment of roles. In case A) DDQN is the attacker and N2D is the defender, but in case B) DDQN becomes the defender and N2D becomes the attacker. The attacking agent has access to 32 action outputs and the defending agent has access to 68 action outputs. The attacker has one action, selecting a host to be hacked and compromising it, but with 32 hosts, we have 32 available action outputs. The defender has 4 different types of actions; (1) Isolating and patching a host; (2) Reconnecting a host and its respective links; (3) Migrating the critical server to any of the predetermined backup locations (hosts) and; (4) doing nothing. The single action rewards are within the set $R = \{-1, 0, 1\}$ such that $r \in R$, however for the attacker the action rewards are $r \in \{-1, 1\}.$

\section{RESULTS AND DISCUSSIONS}
The results are presented in two categories: game performance results and agent rewards. These results are presented in the next subsections.

\subsection{Game Results}
Table \ref{Table:Game_Results} shows the results of the two games played. 10 Runs with each agent having 25,000 turns was done. The number of turns taken for the defending agent to win each game recorded, and statistically analysed. Statistical analysis showed the two-tailed p-value equals 0.1449, revealing there is no significant statistical difference between the two algorithms. The calculated mean value for the number of turns taken for the defending agent using the N2D algorithm is 4,159.20. This implies that on average, it takes the defender 4,159 turns to isolate the intruder, and win the game. The calculated mean value for the defending agent using the DDQN algorithm is 1,823.80. This implies that on average it takes the defender 1,823 turns to isolate the intruder, and win the game. Even though, this shows that the better defender algorithm seems to be DDQN, statistically, there is no significant difference between the algorithms. The large disparity between the means is likely due to some outlier data effects. 
\begin{table}[h]
\centering
  \caption{Game Run Results}\label{Table:Game_Results}
\begin{tabular}{P{1cm}P{3cm}P{3cm}}
\hline
\textbf{Runs} & \textbf{Game 1: total turns played by defender} & \textbf{Game 2: total turns played by defender}\\
 \hline
 1  & 168 & 35\\
 2  & 356 & 68\\
 3  & 3066 & 240\\
 4  & 11120 & 5414\\
 5  & 10258 & 1158\\
 6  & 6154 & 2299\\
 7  & 8726 & 2936\\
 8  & 322 & 4050\\
 9  & 682 & 1699\\
 10  & 740 & 339\\
 \hline
\end{tabular}
\end{table}

In order to find out if there was a significant difference in the means, a two-tailed T-test analysis that was conducted on the two game results presented in Table \ref{Table:Game_Results}, with an alpha of 0.05. Results in Table \ref{Table:Game_Results_Ttest} show a P($T \leq t$) two-tail value is 0.086288326, which is greater than 0.05. Therefore, there is no significant difference between the two algorithms. The DDQN is, therefore, a better algorithm simply because it is simple and leads to lower computational overhead.   

\begin{table}[h]
\centering
  \caption{Two-tailed T-test Analysis for the two games}\label{Table:Game_Results_Ttest}
\begin{tabular}{P{3cm}P{2cm}P{2cm}}
\hline
\textbf{ } & \textbf{Game 1 turns} & \textbf{Game 2 turns}\\
 \hline
Mean & 4159.2 & 1823.8\\
Variance  & 20064225.96 & 3418095.95\\
Observations & 10 & 10\\
Pearson Correlation  & 0.53 & \\
Hypothesized Mean Difference & 0 \\
df & 9 \\
t Stat & 1.9255316 \\
P($T \leq t$) one-tail & 0.04314416 \\
t Critical one-tail & 1.8331129 \\
P($T \leq t$) two-tail & 0.086288326\\
t Critical two-tail & 2.262157158218\\ 
 \hline
\end{tabular}
\end{table}

\subsection{Agent Rewards}
The agents need to learn while being deployed, therefore they need to maximize their received reward, since they may not reach the state where they no longer need to do exploration \cite{poole_mackworth_2018}. Figure 4 shows the rewards distribution of each agent's algorithm in each game’s run. The defending agent uses the N2D algorithm and the attacking agent uses the DDQN algorithm. 
\begin{figure}[h]
    \centering
    \includegraphics[scale=0.4]{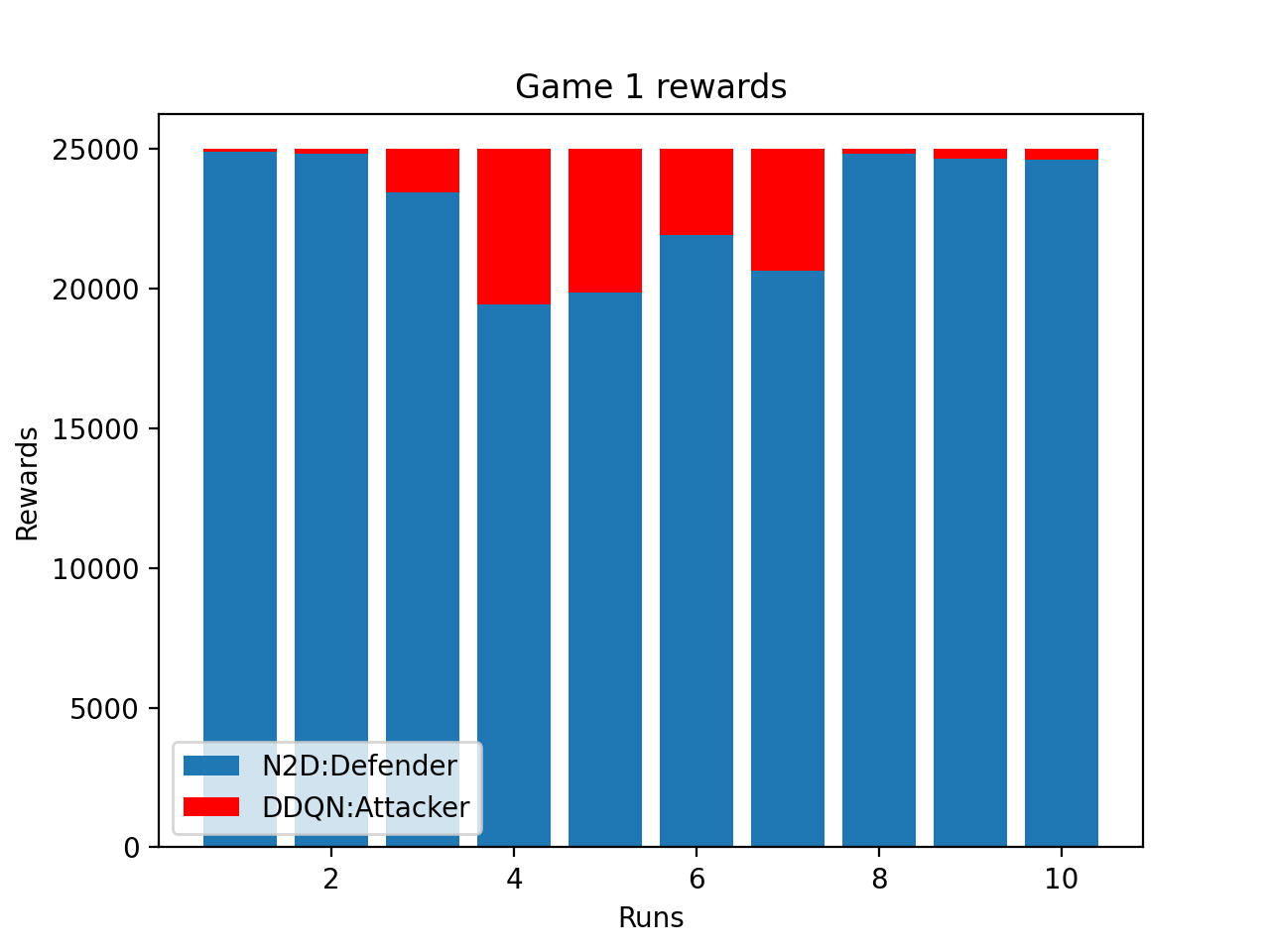}
    \caption{Game run rewards of the defending agent using the neural episodic control to deep q-network algorithm, and attacking agent using the double deep q-network algorithm.}
    \label{fig:mesh1}
\end{figure}
\begin{figure}[h]
    \centering
    \includegraphics[scale=0.4]{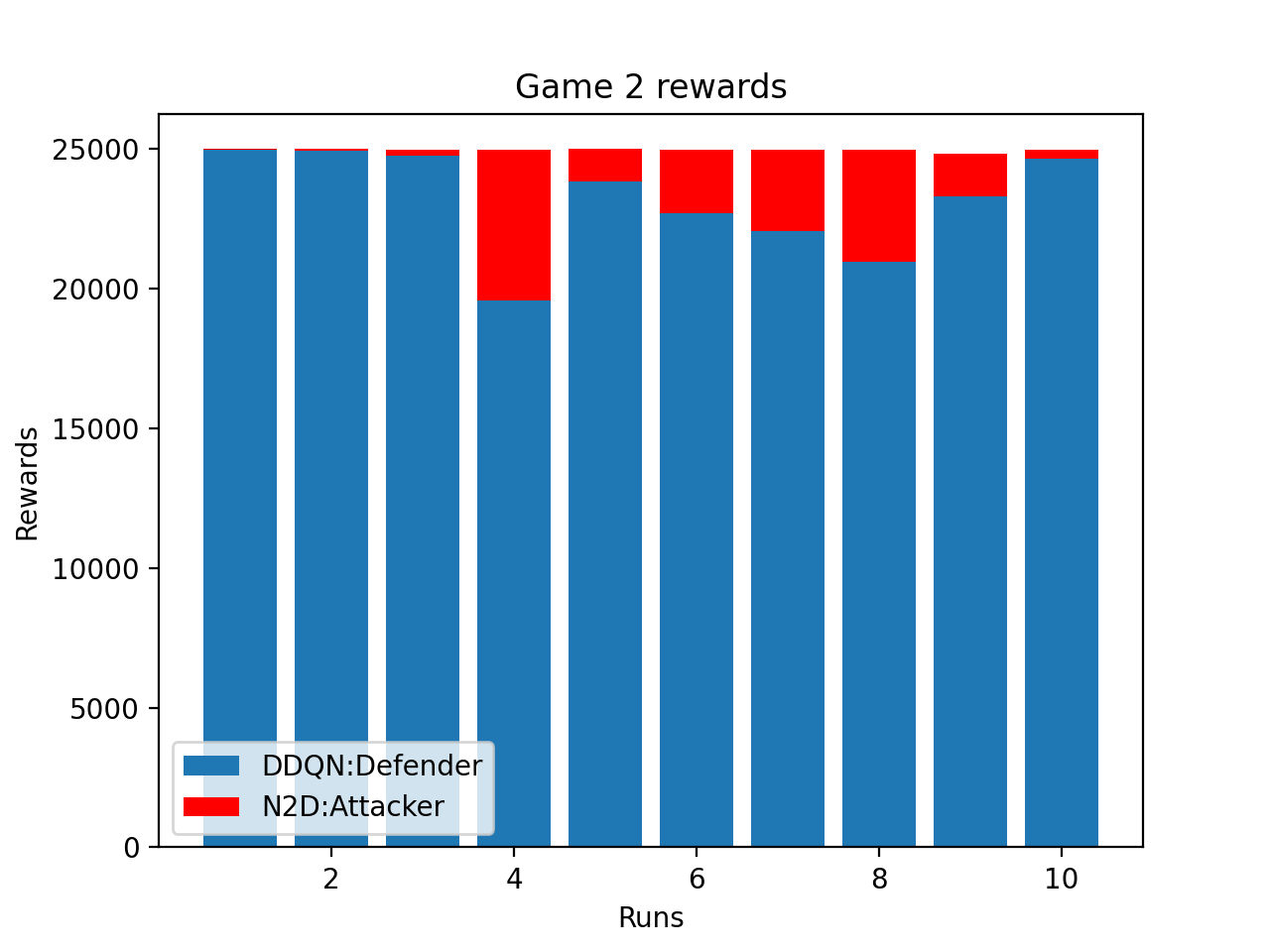}
    \caption{Game run rewards of the attacking agent using the neural episodic control to deep q-network algorithm and the attacking agent using the double deep q-network algorithm.}
    \label{fig:mesh1}
\end{figure}

Figure 5 shows the rewards distribution of each agent's algorithm in each game’s run. Defending agent uses the DDQN algorithm and the attacking agent uses the N2D algorithm. In both games' results the defending agent’s scores in the early and latter runs shows dominance over the attacking agent. This is likely due to the case of exploration being greater than exploitation, and with the defending agent being able to isolate any host, this gives it a much better initial momentum. Unlike the attacker that is limited to the neighbours of already compromised hosts. 
In the mid section group of runs, the attacking agent is able to gain ground, this is likely because at this point the agent has switched to exploitation and is able to target hosts more efficiently, leading to longer in-game engagement and a higher turn count. 
In the latter grouping, the defending agent once again dominates, this could be from exploitation and the adaptation to the selection patterns used by the attacker. It is also worth noting that each agent retains its memory of each run. 

\section{CONCLUSION AND FUTURE}
This paper presents a comparative evaluation of Neural Episodic Control to Deep Q-Network (N2D) and Double Deep Q-Networks for cyber security purposes in software defined networks (SDNs). The results show that both algorithms are adequate tools for network defense. There is no statistically significant difference between the approaches, which makes DDQN more favorable due to its simplicity. 

In future, the number of game runs will be increased in order to provide a greater amount of total steps for each agent, as well as a greater data pool to analyse. Adversarial learning will be implemented, in which the attacker does white and black box causative attacks on the defending agent. The network topology will also be expanded. Another issue is that improvement on these algorithms holds the possibility of diminishing returns as a defender but significant growth as an attacker. The current game environment is inherently biased in favor of the defender, therefore improving on the algorithms to function better in this environment also opens the door for the models to become more effective in aggression as an attacker. Should there be growth in the attacking role it should be considered that these algorithm may also serve as good offensive tools within the cyber security space.

\begin{IEEEbiographynophoto}{Luke D. Borchjes}
received his Bachelor's Degree (Honors) in Computer Science from the University of the Western Cape. He is currently working on his MSc in Computer Science at the University of the Western Cape. His research
interests are in Deep Reinforcement Learning, Cyber Security and Software Defined Networks.
\end{IEEEbiographynophoto}

\vspace{-1.0cm}

\begin{IEEEbiographynophoto}{Clement N. Nyirenda}
received his PhD in Computational Intelligence from Tokyo Institute of Technology in 2011. His research interests are in Computational Intelligence paradigms such as Fuzzy Logic, Swarm Intelligence, and Artificial Neural Networks and their applications in Communications.
\end{IEEEbiographynophoto}

\vspace{-1.0cm}

\begin{IEEEbiographynophoto}{Louise Leenen}
 Louise Leenen completed her PhD at the University of Wollongong in Australia in 2009. Her research areas are AI Applications in Cybersecurity, Ontology Engineering and Mathematical Modelling.
\end{IEEEbiographynophoto}

\vfill

\end{document}